\documentclass[runningheads]{llncs}
\usepackage[T1]{fontenc}

\usepackage{color,array}

\usepackage{graphicx}

\usepackage{amsmath}

\usepackage{textcomp}
\usepackage{stfloats}
\usepackage{url}
\usepackage{hyperref}
\usepackage{verbatim}
\usepackage{graphicx}
\usepackage{xcolor}
\usepackage{multirow}
\usepackage{graphicx,xfp}
\usepackage{booktabs,threeparttable}
\usepackage{tikz}
\usepackage{tikzscale}
\usepackage{comment}

\usepackage{caption}
\usepackage{subcaption}

\usepackage{algorithm}
\usepackage{algpseudocode}

\usepackage{pgfplots}
\usepgfplotslibrary{groupplots}
\usepackage{tikz}
\usepackage{tikzscale}
\usetikzlibrary{positioning}

\begin{document}

\title{Beyond Accuracy: Measuring Representation Capacity of Embeddings to Preserve Structural and Contextual Information}

\author{Sarwan Ali
\\
sali85@student.gsu.edu
\authorrunning{S. Ali}
%
\institute{Georgia State University, Atlanta GA 30303, USA }
}



\maketitle

\begin{abstract}
Effective representation of data is crucial in various machine learning tasks, as it captures the underlying structure and context of the data. Embeddings have emerged as a powerful technique for data representation, but evaluating their quality and capacity to preserve structural and contextual information remains a challenge. In this paper, we address this need by proposing a method to measure the \textit{representation capacity} of embeddings. The motivation behind this work stems from the importance of understanding the strengths and limitations of embeddings, enabling researchers and practitioners to make informed decisions in selecting appropriate embedding models for their specific applications. By combining extrinsic evaluation methods, such as classification and clustering, with t-SNE-based neighborhood analysis, such as neighborhood agreement and trustworthiness, we provide a comprehensive assessment of the representation capacity. Additionally, the use of optimization techniques (bayesian optimization) for weight optimization (for classification, clustering, neighborhood agreement, and trustworthiness) ensures an objective and data-driven approach in selecting the optimal combination of metrics. The proposed method not only contributes to advancing the field of embedding evaluation but also empowers researchers and practitioners with a quantitative measure to assess the effectiveness of embeddings in capturing structural and contextual information. For the evaluation, we use $3$ real-world biological sequence (proteins and nucleotide) datasets and performed representation capacity analysis of $4$ embedding methods from the literature, namely Spike2Vec, Spaced $k$-mers, PWM2Vec, and AutoEncoder. Experimental results demonstrate the Spaced $k$-mers-based embedding shows better representation capacity on $2$ out of three datasets. From the weights computed through optimization, we observed that classification, clustering, and trustworthiness hold the maximum weights while neighborhood agreement weight share towards the representation capacity score is very small.
The first of its kind study in the domain of bioinformatics (to the best of our knowledge), the efficacy of the proposed method in accurately measuring the representation capacity of embeddings will lead to improved decision-making and performance in various machine-learning applications in healthcare.
\end{abstract}

\keywords{
Embeddings \and Data representation \and Representation capacity \and t-SNE \and Optimization
}

\section{Introduction}

Effective data representation is vital in various machine learning tasks as it captures the underlying structure and context of the data~\cite{najafabadi2015deep}. It enables accurate modeling and decision-making~\cite{runck2019using}.
Embeddings, as low-dimensional vector representations, have gained prominence for data representation due to their ability to capture meaningful relationships and semantic information in the data~\cite{banerjee2017intelligent}.

While embeddings are widely used, assessing their quality and capacity to preserve structural and contextual information is challenging. Existing evaluation methods often focus on specific aspects and fail to provide a comprehensive assessment of representation capacity.
Current evaluation methods may lack holistic evaluation criteria, focusing on individual tasks or aspects of embeddings. This leads to a limited understanding of their overall effectiveness.
Knowing the strengths and limitations of embeddings is crucial for selecting appropriate models, improving performance, and avoiding unintended biases or inaccuracies in downstream applications~\cite{glielmo2021unsupervised}.
The quality of embeddings directly affects the performance and reliability of machine learning systems, making it essential to have reliable metrics to assess representation capacity~\cite{bian2014knowledge}.

The primary objective of this research is to develop a robust and comprehensive evaluation framework to measure the representation capacity of embeddings.
The proposed framework combines extrinsic evaluation methods, such as classification and clustering, with neighborhood analysis~\cite{chourasia2022informative} and trustworthiness~\cite{pandey2016trustworthiness} using t-SNE to provide a holistic assessment.
By integrating classification and clustering tasks with t-SNE-based neighborhood analysis and trustworthiness, we capture both functional and structural aspects of the embeddings.
Prior research has proposed various evaluation techniques, including intrinsic evaluation measures like word similarity and analogy tasks~\cite{pandey2016trustworthiness}, as well as extrinsic evaluation through downstream tasks~\cite{ali2021k,ali2021spike2vec}. However, current evaluation methods often lack a comprehensive evaluation framework that combines multiple metrics and fails to address the full representation capacity of embeddings.
The proposed method overcomes the limitations of existing approaches by providing a comprehensive assessment of representation capacity, considering both functional and structural aspects.

An alternative approach to embedding design in sequence classification is the utilization of a kernel (gram) matrix. Kernel-based machine learning classifiers, such as Support Vector Machines (SVM)~\cite{farhan2017efficient}, can leverage kernel matrices for effective classification. These methods have shown promising results compared to feature engineering-based techniques~\cite{ali2021k}. In this approach, the kernel matrix is computed by evaluating the similarity (kernel) values between sequences based on the number of matches and mismatches between $k$-mers~\cite{mismatchProteinClassification}. The resulting kernel matrix can be employed not only with kernel-based classifiers like SVM but also with non-kernel-based classifiers such as decision trees using kernel Principal Component Analysis (PCA)~\cite{hoffmann2007kernel}. However, the kernel-based approach faces two main challenges:
\begin{itemize}
    \item \textbf{Computation of Pairwise Sequence Similarity:} Computing the pairwise sequence similarity required for constructing the kernel matrix can be computationally expensive. As the number of sequences increases, the computational cost grows significantly, limiting the scalability of the kernel-based method.
    \item \textbf{Memory Storage of Large Kernel Matrices:} Storing a kernel matrix of dimensions $n \times n$, where $n$ represents the number of sequences, can be challenging, especially when $n$ is very large. The memory requirements for such matrices can become prohibitive, making it difficult to scale the kernel-based method to handle a large number of sequences effectively.
\end{itemize}

Due to the problems discussed above for kernel-based methods, feature engineering and deep learning-based methods for embedding design are more popular among researchers. For this purpose, we only focus on those types of embeddings in this paper. The proposed evaluation framework incorporates classification tasks to assess the discriminative power of embeddings and clustering analysis to evaluate their ability to capture inherent data clusters. The neighborhood structures of embeddings in high-dimensional and low-dimensional spaces are compared using t-SNE to evaluate the preservation of local relationships. The Bayesian optimization approach is employed to optimize the weights assigned to different evaluation metrics including classification, clustering, neighborhood agreement, and trustworthiness, ensuring a balanced assessment of representation capacity.
The proposed method offers a comprehensive and holistic understanding of embedding quality. The incorporation of an optimization approach, which allows for the automatic selection of weights, ensures an objective and data-driven approach to measuring the importance of different evaluation metrics. The proposed method equips researchers and practitioners with a robust and quantifiable measure to assess the effectiveness of embeddings in preserving structural and contextual information, enabling informed decision-making in selecting appropriate embedding models.
Our contributions to this paper are summarized as follows:
\begin{enumerate}
    \item \textbf{Development of a Comprehensive Evaluation Framework:} We propose a novel evaluation framework to measure the representation capacity of embeddings. Unlike existing methods that focus on specific aspects, our framework integrates classification, clustering, t-SNE-based neighborhood analysis, and trustworthiness to provide a holistic assessment. This comprehensive approach enables a thorough understanding of the effectiveness of embeddings in preserving structural and contextual information.
    \item \textbf{Incorporation of Optimization Technique:} To ensure an objective and data-driven evaluation, we employ Bayesian optimization for weight optimization. This approach automatically selects optimal weights for different evaluation metrics, including classification, clustering, neighborhood agreement, and trustworthiness. By optimizing the weights, we achieve a balanced assessment of representation capacity, considering the relative importance of each metric.
    \item \textbf{Application to Real-World Biological Sequence Datasets:} We apply our evaluation framework to three real-world biological sequence datasets, including proteins and nucleotides. By analyzing the representation capacity of four embedding methods from the literature (Spike2Vec, Spaced $k$-mers, PWM2Vec, and AutoEncoder), we demonstrate the practicality and effectiveness of our approach in assessing different embedding models.
    \item \textbf{Identification of Embedding Strengths and Weaknesses:} Through the evaluation process, we identify the strengths and weaknesses of the evaluated embedding methods. Specifically, we observe that Spaced $k$-mers-based embedding shows superior representation capacity on two out of three datasets. This insight provides valuable guidance for researchers and practitioners in selecting the most suitable embedding models for specific tasks.
    \item \textbf{Novelty in the Domain of Bioinformatics:} To the best of our knowledge, this study represents the first comprehensive evaluation of representation capacity in the domain of bioinformatics. By quantitatively measuring the quality of embeddings, our research contributes to improved decision-making and performance in various machine learning applications in healthcare.
\end{enumerate}

The rest of the paper is organized as follows. The discussion of existing related studies is given in Section~\ref{sec_related_work}. The proposed method to compute the representation capacity score is given in Section~\ref{sec_proposed_approach}. The detail regarding the experimental setup and dataset is reported in Section~\ref{sec_exp_setup}. We report the findings from our research in Section~\ref{sec_RD}.
Finally, the paper is concluded in Section~\ref{sec_conclusion}.

\section{Related Work}\label{sec_related_work}
Various evaluation methods have been proposed to assess the quality and effectiveness of embeddings in capturing data semantics and relationships~\cite{zhang2018metagraph2vec,wang2019evaluating,schnabel2015evaluation}. Intrinsic evaluation measures, such as word similarity~\cite{antoniak2018evaluating} and analogy tasks~\cite{hartmann2017portuguese}, evaluate embeddings based on their ability to capture linguistic properties and semantic similarities~\cite{wang2018comparison}. These measures provide insights into the semantic representation capabilities of embeddings~\cite{shor2020towards,wang2018comparison}.

Extrinsic evaluation approaches assess the performance of embeddings in downstream tasks, such as sentiment analysis~\cite{yu2017refining}, named entity recognition~\cite{akbik2019pooled}, and machine translation~\cite{zou2013bilingual}. These evaluations measure the impact of embeddings on task-specific performance, providing a practical evaluation of their usefulness in real-world applications~\cite{wang2022survey,bhatia2015sparse,rudkowsky2018more}.

Existing evaluation methods for embeddings suffer from several limitations and gaps that hinder a comprehensive assessment of their representation capacity~\cite{ben2002limitations}. Firstly, many methods focus on specific aspects of embeddings, such as semantic similarity~\cite{plank2013embedding} or performance on individual tasks~\cite{wang2021position,nayak2016evaluating}, without considering the broader context of representation capacity.

Secondly, the evaluation metrics often lack a holistic approach, failing to capture the full spectrum of functional and structural aspects of embeddings~\cite{beckman2007innovation,yu2014deep}. This limited perspective can lead to incomplete assessments and overlook potential weaknesses or biases in the embeddings.

Several authors proposed feature engineering~\cite{ali2021k,ali2022pwm2vec,kuzmin2020machine} and deep learning-based methods~\cite{ali2022spike2signal} to design embeddings for different downstream tasks, such as classification and clustering~\cite{tayebi2021robust}.
Authors in~\cite{wang2017time} use the ResNet model to perform classification.
However, in the case of tabular data, the deep learning models show suboptimal results~\cite{shwartz2022tabular}.
Authors in~\cite{lochel2021chaos,lochel2020deep} convert the biological sequences to images, which can then be applied for classification using deep learning image classifiers. However, transforming a biological sequence into an image without loss of information is still a challenging task. Moreover, it is not known how much information is preserved in the images.

The proposed method fills the gaps in current evaluation approaches by providing a comprehensive assessment of representation capacity. By combining multiple evaluation metrics, including classification, clustering, and neighborhood analysis, the proposed framework captures both the discriminative power and the preservation of structural and contextual information.

\section{Proposed Approach}\label{sec_proposed_approach}

In this section, we present our proposed methodology for evaluating the representation capacity of embeddings. Our approach combines classification and clustering evaluation with neighborhood analysis using t-SNE and leverages Bayesian optimization approach for weight optimization.

\subsection{Embedding Generation}
Given the biological sequences as input, the first step is to generate fixed-length numerical representation from variable-length sequences. For this purpose, we use the idea of an embedding method, called Spike2Vec, proposed in~\cite{ali2021spike2vec}.

\subsubsection{Spike2Vec}
Given a biological sequence $s$ as input, this method first generates substrings (called mers) of length $k$ (hence $k$-mers). For example, if the sequence is ``ATCGGCA" and k=3, the k-mers would be "ATC", "TCG", "CGG", "GGC", "GCA". We generate all possible k-mers for the entire sequence. 
The total number of $k$-mers that can be generated from a given sequence are:
\begin{equation}
    \vert k-mers \vert = \vert s \vert - k + 1
\end{equation}
where $\vert s \vert$ is the length of the biological sequence.
Generating $k$-mers basically means breaking down the original sequence into overlapping k-mers. Note that the value of $k$ is a tunable parameter, which is selected using the standard validation set approach~\cite{validationSetApproach}.

The next step is to count the frequency of occurrence for each $k$-mer in the sequence. This will create a $k$-mer spectrum, which represents the distribution of k-mers in the sequence. The length of the $k$-mer spectrum equals all possible $k$-mers within a sequence. Formally, given the alphabet $\Sigma$ (where $\Sigma$ corresponds to an alphabet comprised of all possible characters within a biological sequence \textit{ACDEFGHIKLMNPQRSTVWXY}), the length of a spectrum is $\vert \Sigma \vert^k$, which contains the count of $k$-mers within a sequence. For this method, we took $k=3$. We then normalize the $k$-mer spectrum to account for variations in sequence length. For this purpose, we divide the count of each $k$-mer by the total number of $k$-mers to obtain the normalized frequency, which we call Spike2Vec, which captures the information about the distribution and frequency of $k$-mers in the sequence.

\subsubsection{Spaced $k$-mers}
The feature vectors generated based on the frequencies of $k$-mers in sequences tend to be large and sparse, which can have a detrimental effect on sequence classification performance. To mitigate this issue, the concept of spaced $k$-mers was introduced~\cite{singh2017gakco}, aiming to generate compact feature vectors with reduced sparsity and size. Spaced $k$-mers involve using non-contiguous length $k$ subsequences, referred to as $g$-mers. Given a biological sequence as input, the algorithm first computes $g$-mers and then derives $k$-mers from those $g$-mers, where the value of $k$ is less than $g$. In our experiments, we specifically used $k=4$ and $g=9$. The size of the gap between consecutive $k$-mers is determined by $g-k$. The resultant $k$-mers are then used to compute the spectrum of length $\vert \Sigma \vert^k$ as done in the case of Spike2Vec.

\subsubsection{PWM2Vec}
The Spike2Vec method yields frequency vectors that are relatively low-dimensional but still maintain a high-dimensional representation. However, the process of matching $k$-mers to their corresponding location/bin in the vector (bin matching) can be computationally intensive. To address these challenges, PWM2Vec~\cite{ali2022pwm2vec} has been introduced as a potential solution. PWM2Vec leverages the concept of position-weight matrices (PWMs) to generate a fixed-length numerical feature vector. By constructing a PWM from the $k$-mers in the sequence, PWM2Vec assigns a score to each $k$-mer in the PWM and incorporates both localization information and the significance of each amino acid's position in the sequence. This approach allows for the creation of a concise and comprehensive feature embedding that can be applied to various machine learning tasks downstream. PWM2Vec offers a more efficient and effective alternative to the computation of $k$-mer frequency vectors, combining important information in a compact manner.

\subsubsection{AutoEncoder}
The AutoEncoder-based approach~\cite{xie2016unsupervised} uses a deep neural network to learn a compact feature representation of the input data. This is achieved by employing a non-linear mapping technique that transforms the data space $X$ into a lower-dimensional feature space $Z$. Given one-hot encoding-based vectors as input, this approach iteratively optimizes an objective function to refine the feature representation. In our experiments, we employed a two-layered neural network with an ADAM optimizer and Mean Squared Error (MSE) loss function. The sequences serve as the input to the network, and through the training process, the network learns to extract meaningful and discriminative features from the data.

\subsection{Classification and Clustering Evaluation}

To assess the discriminative power and clustering ability of embeddings, we perform both classification and clustering tasks.

\subsubsection{Classification Evaluation}
We train a classification model, called logistic regression classifier, using the Spike2Vec embedding as input features. The classification accuracy, denoted as $Acc_{class}$, measures the model's ability to correctly classify instances based on the embeddings.

\subsubsection{Clustering Evaluation}
We apply clustering algorithms, called k-means to group instances based on the embeddings. The number of clusters selected for each dataset equals the number of classes within the dataset (as tager labels are available in the datasets used for experimentation).
We evaluate the quality of clustering using the silhouette score, denoted as $Score_{clust}$. A higher $Score_{clust}$ indicates better clustering performance.

\subsubsection{Silhouette Score}
The silhouette score is a measure of how well each data point in a cluster is separated from points in other clusters. It takes into account both the average distance between points within a cluster (cohesion) and the average distance between points in different clusters (separation). The silhouette score for a data point $i$ can be calculated using the following equation:
\begin{equation}
    Score_{clust}(i) = \frac{b(i) - a(i)}{max(a(i), b(i))}
\end{equation}
where $Score_{clust}(i)$ is the silhouette score for data point $i$, the $a(i)$ is the average distance between data point $i$ and all other points within the same cluster, and $b(i)$ is the average distance between data point $i$ and all points in the nearest neighboring cluster (the cluster that gives the smallest value of b(i)).
The silhouette score ranges from -1 to +1, where a higher value indicates that the data point is well-matched to its own cluster and poorly matched to neighboring clusters. A score close to 1 implies a well-clustered data point, while a score close to -1 suggests that the data point may be assigned to the wrong cluster. A score around 0 indicates that the data point is on or very close to the decision boundary between two neighboring clusters. The average silhouette score for all data points in a clustering solution provides an overall measure of the quality of the clustering result.

\subsection{Neighborhood Analysis using t-SNE}
To analyze the preservation of neighborhood structures in the embeddings, we employ t-stochastic neighborhood embedding (t-SNE)~\cite{van2008visualizing}, a dimensionality reduction technique that maps high-dimensional embeddings to a low-dimensional space while preserving local relationships. The pseudocode to compute t-SNE is given in Algorithm~\ref{algo_tsne}.

\begin{algorithm}[h!]
\caption{t-SNE Computation}
\label{algo_tsne}
\begin{algorithmic}[1]
\Require High-dimensional data points $X$, perplexity $Perp$, number of iterations $Iter$
\State Compute pairwise Euclidean distances $D$ between data points
\State Initialize low-dimensional embedding $Y$ randomly
\For{$t = 1$ \textbf{to} $Iter$}
\State Compute similarity matrix $P$ using the Gaussian kernel with adaptive perplexity
\State Compute perplexity-based probabilities $Q$ using binary search
\State Compute gradient $\frac{\partial C}{\partial Y}$ using Equation~\ref{eq_gradient}
\State Update low-dimensional embedding $Y$ using gradient descent with momentum
\EndFor
\State \textbf{return} Low-dimensional embedding $Y$
\end{algorithmic}
\end{algorithm}

\subsubsection{Compute Gradient}
An important step in t-SNE is the computation of gradient.
The gradient $\frac{\partial C}{\partial Y}$ in the t-SNE algorithm is computed to optimize the embedding space. It represents the direction and magnitude of the change that needs to be made to the embedding coordinates in order to minimize the cost function $C$. The equation for computing the gradient is as follows:

\begin{equation}\label{eq_gradient}
\frac{\partial C}{\partial Y} = 4 \sum_{i=1}^{N} \left(\sum_{j=1}^{N} P_{ij} - Q_{ij}\right) (Y_i - Y_j)
\end{equation}

where $N$ is the number of data points, $P_{ij}$ is the similarity between points $i$ and $j$ based on the gaussian kernel, $Q_{ij}$ is the perplexity-based probability between points $i$ and $j$, and $Y_i$ and $Y_j$ are the low-dimensional coordinates of points $i$ and $j$.
By computing the gradient using Equation~\ref{eq_gradient} and updating the embedding coordinates accordingly, the t-SNE algorithm optimizes the embedding space to better represent the underlying structure of the high-dimensional data.

Given the original high-dimensional embeddings $X$ and the corresponding t-SNE embeddings $Y$, we calculate the pairwise Euclidean distances between instances in both spaces as $D_X$ and $D_Y$, respectively. The neighborhood agreement, denoted as $Agree_{neighbor}$, is computed using the K-nearest neighbors (KNN) approach:

\begin{equation}
    Agree_{neighbor} = \frac{1}{N} \sum_{i=1}^N \frac{1}{K} \sum_{j \in KNN_i} \delta (\vert \vert X_i - X_j \vert \vert - \vert \vert Y_i - Y_j \vert \vert)
\end{equation}

where $N$ is the total number of instances, $K$ is the number of nearest neighbors considered, and $\delta(\cdot)$ is the indicator function that returns 1 if the argument is true and 0 otherwise. The value for $Agree_{neighbor}$ ranges from $0$ to $100$, where the maximum value is better.
We measure the neighborhood agreement from $K=1$ until $K=100$, normalize them between $0$ and $1$ (using min-max normalization), and take the average to get a scaler value.

\subsection{Trustworthiness Analysis using t-SNE}
To perform a further evaluation using t-SNE, we use trustworthiness, which is a measure that quantifies the extent to which the local relationships among data points are preserved in the embedding space produced by t-SNE. It is computed by comparing the distances between data points in the original high-dimensional space with the distances between the corresponding points in the t-SNE embedding space. The trustworthiness for a given value of $K$ (i.e. number of nearest neighbors), denoted as $Trust(K)$, is calculated using the following equation:
\begin{equation}
Trust(K) = 1 - (\frac{2}{n \times (n - 1)}) \times \sum_{i=1}^{n} \sum_{j=1}^{K} (rank(i, j) - K)
\end{equation}

where $n$ is the total number of data points, $rank(i, j)$ represents the rank of the $j-th$ nearest neighbor of the $i-th$ data point in the original high-dimensional space, and $K$ is the number of nearest neighbors used for the comparison. This process is repeated for different values of $K$ from 1 to 100 to get $Trust_{neighbor}$.

The trustworthiness calculation involves computing the rank of each data point's neighbors in the original space and comparing it with the ranks of the corresponding points' neighbors in the t-SNE embedding space. A lower rank indicates that the corresponding points in the embedding space are closer to each other, implying a higher level of trustworthiness.

By evaluating trustworthiness for various values of $K$ (i.e. various numbers of neighbors ranging from 1 to 100), it is possible to analyze how well the t-SNE algorithm preserves the local relationships of the data (its value ranges from 0 to 1). Higher values of Trust(K) indicate that the local structures are well-maintained, implying higher trustworthiness of the t-SNE embedding. This measure is valuable for assessing the reliability and quality of the t-SNE visualization or clustering results. We measured the trustworthiness from $K=1$ until $K=100$ and took the average to get a scaler value.

\subsection{Weight Optimization}
After computing $Acc_{class}$, $Score_{clust}$, $Agree_{neighbor}$, and $Trust_{neighbor}$, the final step is to combine these $4$ values to get the representation capacity score.
To obtain a balanced evaluation and ensure each metric contributes appropriately to the final representation capacity score, we employ an automatic approach for weight optimization. We aim to find optimal weights for each metric that maximize the overall evaluation performance.

We define the weights as follows: $w_{class}$, $w_{clust}$, $w_{neighb}$, and $w_{trust}$. These weights determine the importance assigned to each metric.

The representation capacity score, denoted as $RC$, is calculated as follows:
\begin{equation}
\begin{aligned}
    RC &= \bigg[ 
    w_{class} \times \frac{Acc_{class}}{max(Acc_{class})} + w_{clust} \times \frac{Score_{clust}}{max(Score_{clust})} \bigg] - 
    \\
    &\bigg[ 
    w_{neighb} \times \frac{Agree_{neighbor}}{max(Agree_{neighbor})} + 
    \\ 
    &w_{trust} \times \frac{Trust_{neighbor}}{max(Trust_{neighbor})}
    \bigg]
\end{aligned}
\end{equation}

where $\max(\cdot)$ represents the maximum value obtained in each metric.

We employ a Bayesian optimization approach to iteratively search for the optimal weights that maximize the representation capacity score. The optimization process is performed using the Optuna framework~\footnote{\url{https://optuna.org/}}.
Optuna is a library for hyperparameter optimization that utilizes Bayesian optimization as its underlying algorithm. Bayesian optimization is a sequential model-based optimization technique that aims to find the global optimum of an objective function by iteratively exploring the search space and updating a probabilistic model of the objective function.

The key idea behind Bayesian optimization is to model the unknown objective function using a surrogate model. This surrogate model approximates the true objective function and provides estimates of its values at unexplored points in the search space. The surrogate model is often a probabilistic model such as a Gaussian Process (GP) or a Tree-structured Parzen Estimator (TPE).

The Bayesian optimization process consists of several steps. Initially, a set of hyperparameter configurations is randomly sampled from the search space and evaluated using the objective function. This initial data is used to train the surrogate model.

Next, an acquisition function is defined to guide the search for the next promising hyperparameter configuration. The acquisition function balances exploration and exploitation by considering both the surrogate model's predictions and its uncertainty. Commonly used acquisition functions include Expected Improvement (EI), Probability of Improvement (PI), and Upper Confidence Bound (UCB).

The acquisition function is optimized to find the hyperparameter configuration that maximizes its value. This configuration is then evaluated using the objective function, and the resulting data is used to update the surrogate model. The process iterates, with the surrogate model and acquisition function being updated based on the new data.

The optimization process continues until a stopping criterion is met, such as reaching a maximum number of iterations or the objective function converging to a satisfactory value. In our case, we use the $1000$ number of iterations as the stopping criteria.
The algorithm aims to iteratively explore promising regions of the search space and focus on areas that are likely to yield better results.

In summary, Optuna utilizes Bayesian optimization to efficiently search the hyperparameter space by iteratively updating a surrogate model and selecting promising hyperparameter configurations using an acquisition function. This approach allows for an adaptive and data-driven exploration of the search space, leading to improved performance in hyperparameter optimization tasks.

\section{Experimental Setup}\label{sec_exp_setup}
This section presents the dataset utilized in our experiments along with experimental setting details. The experiments were conducted on a system featuring an Intel Core i5 processor running at $2.40$ GHz, coupled with $32$ GB of memory, and operated on the Windows operating system. 

\subsection{Dataset Statistics}
We use $3$ different biological datasets in this study to compute the embeddings and evaluate their representation capacity. The detail of each dataset is described below.

\subsubsection{Spike7k Dataset}
The Spike7k dataset consists of aligned spike protein sequences obtained from the GISAID database~\footnote{\url{https://www.gisaid.org/}}. The dataset comprises a total of 7000 sequences, which represent 22 different lineages of coronaviruses (class labels). Each sequence in the dataset has a length of 1274 amino acids. The distribution of lineages (class labels) in the Spike7k dataset is the following:
B.1.1.7 (3369), B.1.617.2  (875), AY.4 (593), B.1.2 (333), B.1 (292), B.1.177 (243), P.1 (194), B.1.1 (163), B.1.429 (107), B.1.526 (104), AY.12 (101), B.1.160 (92), B.1.351 (81), B.1.427 (65), B.1.1.214 (64), B.1.1.519 (56), D.2 (55),  B.1.221 (52), B.1.177.21 (47), B.1.258 (46), B.1.243 (36), R.1 (32).

\subsubsection{Protein Subcellular}
The Protein Subcellular dataset~\cite{ProtLoc_website_url} comprises unaligned protein sequences annotated with information on 11 distinct subcellular locations, which are used as class labels for classification tasks. The dataset contains a total of 5959 sequences. 
The classes along with their counts in this dataset are the following:
Cytoplasm (1411), Plasma Membrane (1238), Extracellular Space (843), Nucleus (837), Mitochondrion (510), Chloroplast (449), Endoplasmic Reticulum (198), Peroxisome (157), Golgi Apparatus (150), Lysosomal (103), Vacuole (63).

\subsubsection{Human DNA}
This data comprised a collection of unaligned Human DNA nucleotide sequences, comprising a total of 4,380 sequences~\cite{human_dna_website_url}. Each sequence was composed of nucleotides A, C, G, and T. The dataset included a class label indicating the gene family to which each sequence belonged. There were a total of seven unique gene family labels, namely G Protein Coupled, Tyrosine Kinase, Tyrosine Phosphatase, Synthetase, Synthase, Ion Channel, and Transcription Factor. The objective of the study was to classify the gene family of each DNA sequence.
The dataset exhibited variations in sequence lengths. The maximum, minimum, and average sequence lengths in the dataset were found to be 18,921, 5, and 1,263.59, respectively. These statistics provide insights into the range and distribution of sequence lengths within the dataset. 
The classes along with their counts in this dataset are the following: G Protein Coupled (531), Tyrosine Kinase (534), Tyrosine Phosphatase (349), Synthetase (672), Synthase (711), Ion Channel (240), Transcription Factor (1343).

\section{Results And Discussion}\label{sec_RD}
In this section, we report the representation capacity results for different embedding methods generated for different datasets.

The results for the Spike7k dataset are reported in Table~\ref{tbl_Spike7k_Results}. In terms of representation capacity, we can observe that AutoEncoder-based embedding achieves the highest performance. If we break down the performance for different metrics, the Spike2Vec embedding shows the best classification accuracy, AutoEncoder shows the best clustering performance as well as the best trustworthiness value. Moreover, PWM2Vec shows the highest performance in the case of neighborhood agreements. From the ``Optimal Weights (Performance Values)", we can observe that Classification, Clustering, and Trustworthiness got almost equal weight based on the Bayesian optimization. However, neighborhood agreement god very small weight values. Because of this reason, despite PWM2Vec showing the best performance for neighborhood agreement, its representation capacity score is the lowest among all embedding methods (because of the $0.0003$ weight for the neighborhood agreement). The overall behavior shows that the neighborhood agreement may provide additional insights about local structures but might not carry as much discriminative power in the overall analysis.

\begin{table}[h!]
    \centering
    \resizebox{0.9\textwidth}{!}{
    \begin{tabular}{ccp{2.7cm}p{2.7cm}p{2.7cm}p{2.7cm}}
    \toprule
    \multirow{3}{*}{Embedding} & \multirow{3}{2.3cm}{Representation Capacity} & \multicolumn{4}{c}{Optimal Weights (Performance Values)} \\
    \cmidrule{3-6}
    & & Classification & Clustering & Neighborhood Agreement & Trustworthiness \\
    \midrule \midrule
        \multirow{1}{*}{Spike2Vec} & \multirow{1}{*}{0.7638} & 0.3329 (\textbf{0.8533}) & 0.3326 (0.5447) & 0.0074 (0.8623) & 0.3270 (0.9325)  \\
        \cmidrule{2-6}
        \multirow{1}{*}{Spaced $k$-mers} & \multirow{1}{*}{0.7728} & 0.3332 (0.8471) & 0.3330 (0.5589) & 0.0004 (0.8425) &  0.3332 (0.9141)  \\
        \cmidrule{2-6}
        \multirow{1}{*}{PWM2Vec} & \multirow{1}{*}{0.7629} & 0.3333 (0.8171) & 0.3272 (0.5732) & 0.0003 (\textbf{0.8933}) & 0.3390 (0.8942)  \\
        \cmidrule{2-6}
        \multirow{1}{*}{AutoEncoder} & \multirow{1}{*}{\textbf{0.8190}} & 0.3303 (0.7576) & 0.3301 (\textbf{0.7870}) & 0.0094 (0.8336) & 0.3300 (\textbf{0.9597})  \\
         \bottomrule
    \end{tabular}
    }
    \caption{Representation Capacity results for Spike7k dataset. The best values are shown in bold.}
    \label{tbl_Spike7k_Results}
\end{table}

The results for the Protein Subcellular dataset are reported in Table~\ref{tbl_protein_subcellular_Results}. In terms of representation capacity, we can observe that Spaced $k$mers-based embedding achieves the highest performance. If we break down the performance for different metrics, the Spaced $k$-mers embedding shows the best classification accuracy (with an optimal weight of 0.3472) and Clustering performance (with an optimal weight of 0.3136). 
AutoEncoder shows the best Neighborhood Agreement performance while PWM2Vec shows the highest performance in the case of trustworthiness. Again, from the ``Optimal Weights (Performance Values)", we can observe that Classification, Clustering, and Trustworthiness got almost equal weight based on the Bayesian optimization. However, neighborhood agreement god very small weight values, which shows that although neighborhood agreement may provide additional insights about local structures, however, it might not carry as much discriminative power in the overall analysis.
\begin{table}[h!]
    \centering
    \resizebox{0.9\textwidth}{!}{
    \begin{tabular}{ccp{2.7cm}p{2.7cm}p{2.7cm}p{2.7cm}}
    \toprule
    \multirow{3}{*}{Embedding} & \multirow{3}{2.3cm}{Representation Capacity} & \multicolumn{4}{c}{Optimal Weights (Performance Values)} \\
    \cmidrule{3-6}
    & & Classification & Clustering & Neighborhood Agreement & Trustworthiness \\
    \midrule \midrule
        \multirow{1}{*}{Spike2Vec} & \multirow{1}{*}{0.4010} & 0.3335 (0.5687) & 0.3314 (0.0351) & 0.0020 (0.7570) & 0.3329 (0.6045)  \\
        \cmidrule{2-6}
        \multirow{1}{*}{Spaced $k$-mers} & \multirow{1}{*}{\textbf{0.4603}} & 0.3472 (\textbf{0.6677}) & 0.3136 (\textbf{0.1548}) & 0.0001 (0.6183) & 0.3389 (0.5309)  \\
        \cmidrule{2-6}
        \multirow{1}{*}{PWM2Vec} & \multirow{1}{*}{0.4094} & 0.3292 (0.5151) & 0.3303 (0.0634) & 0.0053 (0.6992) & 0.3350 (\textbf{0.6644})  \\
        \cmidrule{2-6}
        \multirow{1}{*}{AutoEncoder} & \multirow{1}{*}{0.3639} & 0.3425 (0.4485) & 0.3145 (0.0017) & 0.00007 (\textbf{0.7634}) & 0.3427 (0.6122)  \\
         \bottomrule
    \end{tabular}
    }
    \caption{Representation Capacity results for Protein Subcellular dataset. The best values are shown in bold.}
    \label{tbl_protein_subcellular_Results}
\end{table}

The results for the Protein Subcellular dataset are reported in Table~\ref{tbl_human_dna_Results}. In terms of representation capacity, we can observe that Spaced $k$mers-based embedding achieves the highest performance despite achieving the best individual performance for only classification metric. If we breakdown the performance for different metrics, the Spike2Vec embedding shows the best Neighborhood agreement value (with optimal weight of 0.0000006 and neighborhood agreement value of 0.9646) and Trustworthiness value (with optimal weight of 0.3305 and neighborhood agreement value of 0.9732). 
The PWM2Vec shows the best Clustering performance. Here we can observe the same pattern as with the previous two datasets where classification, clustering, and Trustworthiness got almost equal weights while neighborhood agreement contains almost $0$ weight.

\begin{table}[h!]
    \centering
    \resizebox{0.9\textwidth}{!}{
    \begin{tabular}{ccp{2.7cm}p{2.7cm}p{2.9cm}p{2.7cm}}
    \toprule
    \multirow{3}{*}{Embedding} & \multirow{3}{2.3cm}{Representation Capacity} & \multicolumn{4}{c}{Optimal Weights (Performance Values)} \\
    \cmidrule{3-6}
    & & Classification & Clustering & Neighborhood Agreement & Trustworthiness \\
    \midrule \midrule
        \multirow{1}{*}{Spike2Vec} & \multirow{1}{*}{0.6515} & 0.3320 (0.5859) & 0.3373 (0.4009) & 0.0000006 (\textbf{0.9646}) & 0.3305 (\textbf{0.9732})  \\
        \cmidrule{2-6}
        \multirow{1}{*}{Spaced $k$-mers} & \multirow{1}{*}{\textbf{0.6667}} & 0.3329 (\textbf{0.7313}) & 0.3275 (0.3184) & 0.0002 (0.9637) & 0.3392 (0.9408)  \\
        \cmidrule{2-6}
        \multirow{1}{*}{PWM2Vec} & \multirow{1}{*}{0.5985} & 0.3317 (0.3036) & 0.3319 (\textbf{0.9337}) & 0.0103 (0.7935) & 0.3258 (0.6017)  \\
        \cmidrule{2-6}
        \multirow{1}{*}{AutoEncoder} & \multirow{1}{*}{0.5485} & 0.3459 (0.6605) & 0.3079 (0.0497) & 0.0001 (0.9168) & 0.3459 (0.8810)  \\
         \bottomrule
    \end{tabular}
    }
    \caption{Representation Capacity results for Human DNA dataset. The best values are shown in bold.}
    \label{tbl_human_dna_Results}
\end{table}

\section{Conclusion}\label{sec_conclusion}
In this paper, we have presented a comprehensive evaluation framework to measure the representation capacity of embeddings, addressing the need for robust and quantitative assessment methods. By combining classification, clustering, t-SNE-based neighborhood analysis, and trustworthiness, our framework provides a holistic understanding of the effectiveness of embeddings in capturing structural and contextual information. 
To ensure an objective evaluation, we have employed Bayesian optimization for weight optimization, allowing for the automatic selection of optimal weights for different evaluation metrics. This data-driven approach maximizes the representation capacity score and provides a balanced assessment of embeddings. 
Applying our evaluation framework to real-world biological sequence datasets, including proteins and nucleotides, we have analyzed the representation capacity of four embedding methods.
Future work in this area could focus on incorporating additional evaluation metrics or tasks that could provide a more comprehensive assessment of representation capacity. Exploring other intrinsic and extrinsic evaluation measures specific to different domains or applications could further enhance the evaluation process. 

\bibliographystyle{IEEEtran}
\bibliography{references}

\end{document}